\documentclass{article} 
\usepackage{iclr2026_conference,times}

\usepackage{url}
\usepackage{graphicx}
\usepackage{subcaption}
\usepackage{enumitem}
\usepackage{xspace}
\usepackage{amsmath}
\usepackage{multirow}

\usepackage{microtype}
\usepackage{graphicx}
\usepackage{subcaption}
\usepackage{booktabs} 

\usepackage{hyperref}

\usepackage{amsmath}
\usepackage{amssymb}
\usepackage{mathtools}
\usepackage{amsthm}
\usepackage{xspace}
\usepackage{pgfplots}
\pgfplotsset{compat=1.18}
\usetikzlibrary{patterns}
\usepackage{float}
\usepackage{algorithm}
\usepackage{algorithmic}

\usepackage{amsmath,amsfonts,bm}

\def\Figref#1{Figure~\ref{#1}}

\def\eqref#1{equation~\ref{#1}}

\def\1{\bm{1}}

\def\va{{\bm{a}}}
\def\vb{{\bm{b}}}
\def\vc{{\bm{c}}}
\def\vd{{\bm{d}}}

\def\vh{{\bm{h}}}

\def\vl{{\bm{l}}}

\def\vo{{\bm{o}}}

\def\vu{{\bm{u}}}

\def\mB{{\bm{B}}}

\def\mG{{\bm{G}}}
\def\mH{{\bm{H}}}

\def\mW{{\bm{W}}}

\DeclareMathAlphabet{\mathsfit}{\encodingdefault}{\sfdefault}{m}{sl}
\SetMathAlphabet{\mathsfit}{bold}{\encodingdefault}{\sfdefault}{bx}{n}

\usepackage[capitalize,noabbrev]{cleveref}

\theoremstyle{plain}

\theoremstyle{definition}

\theoremstyle{remark}

\newtoggle{comments}
\togglefalse{comments}

\iftoggle{comments}{
    \newcommand{\austin}[1]{\textcolor{orange}{(Austin: #1)}}
    \newcommand{\shafiq}[1]{\textcolor{cyan}{(shafiq: #1)}}
    \newcommand{\nxphi}[1]{\textcolor{red}{(Phi: #1)}}
    \newcommand{\shrey}[1]{\textcolor{red}{(Shrey)}}
}{
    \newcommand{\austin}[1]{}
    \newcommand{\shafiq}[1]{}
    \newcommand{\nxphi}[1]{}
    \newcommand{\shrey}[1]{}
}

\newcommand{\ourmethod}{{LLEP}\xspace}

\crefname{section}{\S}{\S\S}
\crefname{figure}{Fig.}{Figs.}%
\crefname{equation}{Eq.}{Eqns.}%
\crefname{appendix}{App.}{Apps.}%
\crefname{algorithm}{Alg.}{Algs.}%

\title{Least-Loaded Expert Parallelism: Load Balancing An Imbalanced Mixture-of-Experts}

\author{Xuan-Phi Nguyen, Shrey Pandit, Austin Xu, Caiming Xiong, Shafiq Joty \\
Salesforce AI Research\\
\texttt{xnguyen@salesforce.com}
}

\iclrfinalarxiv
\begin{document}

\maketitle

\begin{abstract}
Mixture-of-Experts (MoE) models are typically pre-trained with explicit load-balancing constraints to ensure statistically balanced expert routing. Despite this, we observe that even well-trained MoE models exhibit significantly imbalanced routing. This behavior is arguably natural—and even desirable—as imbalanced routing allows models to concentrate domain-specific knowledge within a subset of experts. Expert parallelism (EP) is designed to scale MoE models by distributing experts across multiple devices, but with a less-discussed assumption of balanced routing. Under extreme imbalance, EP can funnel a disproportionate number of tokens to a small number of experts, leading to compute- and memory-bound failures on overloaded devices during post-training or inference, where explicit load balancing is often inapplicable. We propose \textbf{Least-Loaded Expert Parallelism (\ourmethod)}, a novel EP algorithm that dynamically reroutes excess tokens and associated expert parameters from overloaded devices to underutilized ones. This ensures that all devices complete their workloads within the minimum collective latency while respecting memory constraints.
Across different model scales, \ourmethod achieves up to 5$\times$ speedup and 4$\times$ reduction in peak memory usage compared to standard EP. This enables faster and higher-throughput post-training and inference, with $\sim$1.9$\times$ faster for gpt-oss-120b. We support our method with extensive theoretical analysis and comprehensive empirical evaluations, including ablation studies. These results illuminate key trade-offs and enable a principled framework for hardware-specific hyperparameter tuning to achieve optimal performance.
\end{abstract}

{\centering
    \href{https://github.com/SalesforceAIResearch/LeastLoadedEP}{\texttt{Code: github.com/SalesforceAIResearch/LeastLoadedEP}}\\[0.5em]
}

\begin{figure}[h!]
\centering
\begin{subfigure}[b]{0.33\columnwidth}
\centering
\begin{tikzpicture}
\begin{axis}[
    ybar,
    width=1.1\textwidth,
    height=5cm,
    ylabel={Speedup ($\times$)},
    ylabel style={font=\scriptsize, yshift=-3pt},
    yticklabel style={font=\scriptsize, xshift=2pt},
    symbolic x coords={16, 4, 1},
    xtick=data,
    xlabel={Imbal. Experts},
    xlabel style={font=\scriptsize},
    ymin=0,
    ymax=5.7,
    bar width=3pt,
    enlarge x limits=0.25,
    nodes near coords,
    nodes near coords style={font=\tiny, rotate=90, anchor=west},
    every node near coord/.append style={yshift=0pt},
    clip=false,
    tick label style={font=\scriptsize},
]
\addplot[fill=gray!50, draw=gray!70] coordinates {(16, 1.0) (4, 1.0) (1, 1.0)};
\addplot[fill=teal!60, draw=teal!80] coordinates {(16, 1.5) (4, 1.8) (1, 1.9)};
\addplot[fill=blue!60, draw=blue!80] coordinates {(16, 2.3) (4, 2.6) (1, 2.8)};
\addplot[fill=orange!70, draw=orange!90] coordinates {(16, 3.7) (4, 3.7) (1, 4.2)};
\addplot[fill=red!60, draw=red!80] coordinates {(16, 4.3) (4, 4.2) (1, 4.9)};
\end{axis}
\end{tikzpicture}
\caption{Speedup: \ourmethod vs.\ EP.}
\label{fig:speedup}
\end{subfigure}
\hspace{-0.5em}
\begin{subfigure}[b]{0.33\columnwidth}
\centering
\begin{tikzpicture}
\begin{axis}[
    ybar,
    width=1.2\textwidth,
    height=5cm,
    ylabel={Peak Mem (GB)},
    ylabel style={font=\scriptsize, yshift=-3pt},
    yticklabel style={font=\scriptsize, xshift=2pt},
    xtick={0,2,4},
    xticklabels={16, 4, 1},
    xlabel={Imbal. Experts},
    xlabel style={font=\scriptsize},
    ymin=0,
    ymax=45,
    bar width=1.5pt,
    enlarge x limits=0.25,
    clip=false,
    tick label style={font=\scriptsize},
]
\addplot[fill=gray!30, draw=gray!70, postaction={pattern=north east lines, pattern color=gray!70}] 
    coordinates {(0, 8.57) (2, 8.57) (4, 8.57)};
\addplot[fill=gray!50, draw=gray!70] 
    coordinates {(0, 6.32) (2, 6.32) (4, 6.32)};
\addplot[fill=teal!30, draw=teal!70, postaction={pattern=north east lines, pattern color=teal!70}] 
    coordinates {(0, 15.58) (2, 18.29) (4, 18.88)};
\addplot[fill=teal!60, draw=teal!80] 
    coordinates {(0, 7.50) (2, 6.80) (4, 6.39)};
\addplot[fill=blue!30, draw=blue!70, postaction={pattern=north east lines, pattern color=blue!70}] 
    coordinates {(0, 23.58) (2, 25.52) (4, 25.95)};
\addplot[fill=blue!60, draw=blue!80] 
    coordinates {(0, 8.70) (2, 7.35) (4, 7.05)};
\addplot[fill=orange!30, draw=orange!70, postaction={pattern=north east lines, pattern color=orange!70}] 
    coordinates {(0, 35.59) (2, 36.37) (4, 36.53)};
\addplot[fill=orange!70, draw=orange!90] 
    coordinates {(0, 8.70) (2, 8.16) (4, 8.04)};
\addplot[fill=red!30, draw=red!70, postaction={pattern=north east lines, pattern color=red!70}] 
    coordinates {(0, 41.60) (2, 41.80) (4, 41.82)};
\addplot[fill=red!60, draw=red!80] 
    coordinates {(0, 8.70) (2, 8.57) (4, 8.55)};
\end{axis}
\end{tikzpicture}
\caption{Peak memory (\textbf{lower is better})}
\label{fig:memory}
\end{subfigure}
\hspace{-0.5em}
\begin{subfigure}[b]{0.33\columnwidth}
\centering
\begin{tikzpicture}
\begin{axis}[
    ybar,
    width=1.1\textwidth,
    height=5cm,
    ylabel={Tput (K tok/s)},
    ylabel style={font=\scriptsize, yshift=-3pt},
    yticklabel style={font=\scriptsize, xshift=2pt},
    xtick={1,2,3},
    xticklabels={8, 16, 32},
    xlabel={EP World Size},
    xlabel style={font=\scriptsize},
    ymin=0,
    ymax=400,
    bar width=4pt,
    enlarge x limits=0.25,
    nodes near coords style={font=\tiny, rotate=90, anchor=west},
    clip=false,
    tick label style={font=\scriptsize},
]
\addplot[fill=blue!30, draw=blue!70, postaction={pattern=north east lines, pattern color=blue!70}] 
    coordinates {(1,81.24) (2,146.8) (3,146.8)};
\addplot[fill=blue!60, draw=blue!80, nodes near coords={\pgfmathprintnumber\pgfplotspointmeta$\times$}, 
    point meta=explicit symbolic] 
    coordinates {(1,113.52) [1.4] (2,216.65) [1.5] (3,325.5) [2.2]};
\addplot[fill=orange!30, draw=orange!70, postaction={pattern=north east lines, pattern color=orange!70}] 
    coordinates {(1,80) (2,110.08) (3,131.07)};
\addplot[fill=orange!70, draw=orange!90, nodes near coords={\pgfmathprintnumber\pgfplotspointmeta$\times$}, 
    point meta=explicit symbolic] 
    coordinates {(1,88) [1.1] (2,146.72) [1.3] (3,245.89) [1.9]};
\end{axis}
\end{tikzpicture}
\caption{Full-model throughput}
\label{fig:end_to_end_speedup}
\end{subfigure}

\vspace{0.2em}
\centering
{\scriptsize
\begin{tabular}{@{}r@{\,}l@{\;}l@{\;}l@{\;}l@{\;}l@{\;\;}r@{\,}l@{\;}l@{\;\;}r@{\,}l@{\;}l@{}}
\textbf{Imbal:} & 
\tikz[baseline=-0.5ex]{\fill[gray!50, draw=gray!70] (0,-0.08) rectangle (0.15,0.12);}\,Bal &
\tikz[baseline=-0.5ex]{\fill[teal!60, draw=teal!80] (0,-0.08) rectangle (0.15,0.12);}\,30\% &
\tikz[baseline=-0.5ex]{\fill[blue!60, draw=blue!80] (0,-0.08) rectangle (0.15,0.12);}\,50\% &
\tikz[baseline=-0.5ex]{\fill[orange!70, draw=orange!90] (0,-0.08) rectangle (0.15,0.12);}\,80\% &
\tikz[baseline=-0.5ex]{\fill[red!60, draw=red!80] (0,-0.08) rectangle (0.15,0.12);}\,95\% &
\textbf{gpt-oss:} & 
\tikz[baseline=-0.5ex]{\fill[blue!60, draw=blue!80] (0,-0.08) rectangle (0.15,0.12);}\,20b &
\tikz[baseline=-0.5ex]{\fill[orange!70, draw=orange!90] (0,-0.08) rectangle (0.15,0.12);}\,120b &
\textbf{Method:} & 
\tikz[baseline=-0.5ex]{\fill[black!30, draw=black!70, postaction={pattern=north east lines, pattern color=black!70}] (0,-0.08) rectangle (0.15,0.12);}\,EP &
\tikz[baseline=-0.5ex]{\fill[black!60, draw=black!80] (0,-0.08) rectangle (0.15,0.12);}\,\ourmethod \\
\end{tabular}
}

\caption{
\ourmethod vs.\ standard expert parallelism (EP). \textbf{(a)} \& \textbf{(b)} show the speedup and peak memory usage per GPU of an MoE layer (128 experts, 4 active experts, hidden size of 2048) under perfectly balanced case and various imbalance scenarios: 30\%, 50\%, 80\%, or 95\% of tokens concentrated into 16, 4, 1 imbalanced experts. LLEP is faster than EP by 5$\times$ under extreme imbalance scenarios, while keeping memory usage stable and avoiding out-of-memory risk.
\textbf{(c)} Realistic full-model throughput: up to 2.2$\times$ for gpt-oss-20b and 1.9$\times$ for gpt-oss-120b.
}
\label{fig:speedup_memory_throughput}
\end{figure}
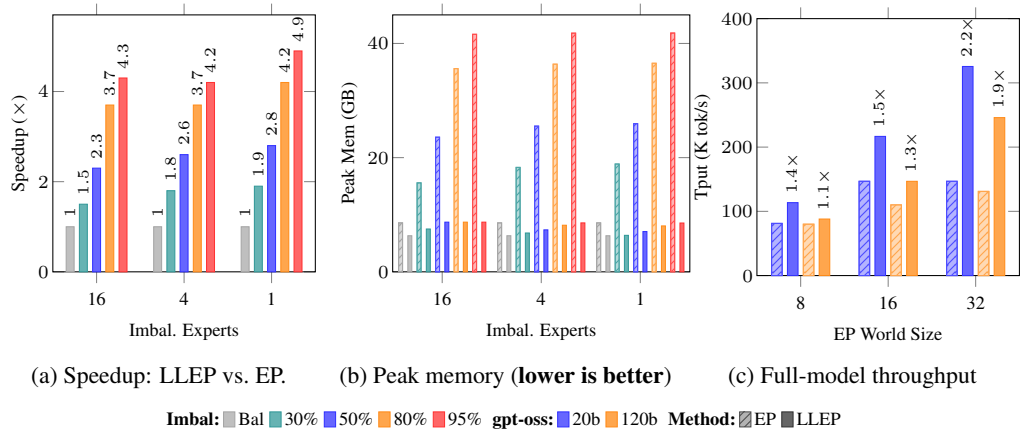

\begin{figure*}[t]
    \centering
    \begin{subfigure}[b]{0.49\textwidth}
        \centering
        \includegraphics[width=\textwidth]{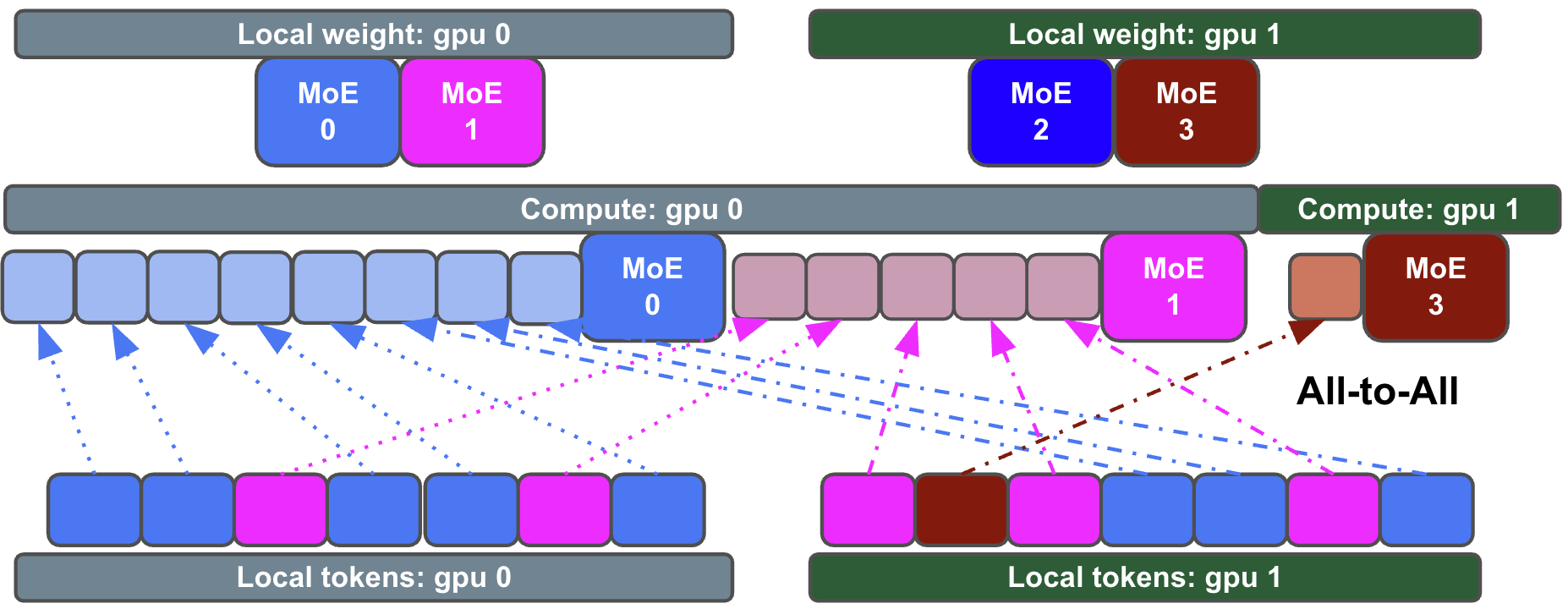}
        \caption{\textbf{Standard EP}: Experts are distributed across devices, but imbalanced routing leads to overloaded devices (gpu 0) and underutilized ones (gpu 1).}
        \label{fig:standard_ep}
    \end{subfigure}
    \hfill
    \begin{subfigure}[b]{0.49\textwidth}
        \centering
        \includegraphics[width=\textwidth]{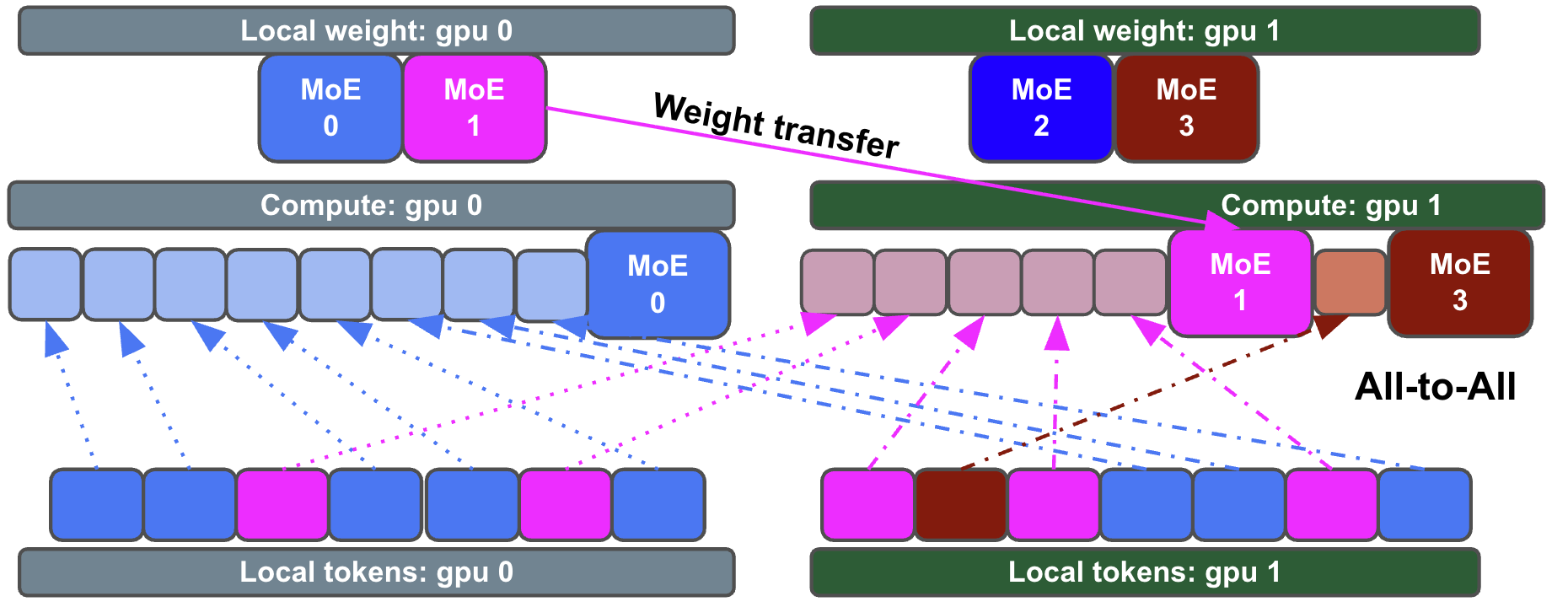}
        \caption{\textbf{\ourmethod}: Dynamically redistributes excess tokens and corresponding expert weights from overloaded devices to underloaded devices for balanced execution.}
        \label{fig:loadbal_ep}
    \end{subfigure}
    \caption{Comparison of standard Expert Parallelism and \ourmethod.}
    \label{fig:method}
\end{figure*}
    
\section{Introduction}\label{sec:intro}



Mixture-of-Experts (MoE) layers have increasingly become indispensable components in large language models (LLMs) due to their ability to scale model size and sparsity while keeping compute usage per token (activated parameters) constant \citep{deepseek_v3_liu2024,gptoss_agarwal2025,yang2025qwen3,kimi_k2_team2025kimi}. An MoE module consists of a \emph{router} that selects the top-$k$ experts to route each token to, and a set of feed-forward experts that process the routed tokens. During pre-training, an MoE layer is often externally load-balanced so that statistically diverse batches of tokens are routed evenly to all experts. This is either enforced with an auxiliary loss \citep{switch_transformers_fedus2022switch} or stochastic bias terms \citep{deepseek_v3_liu2024}. \textbf{Expert parallelism (EP)} has become the default infrastructure setup for MoE model training and inference \citep{hybrid_tp_ep_singh2023hybrid,zheng2024sglang}. It spreads the expert weights across multiple (GPU) devices. Under EP, tokens from different devices are \textit{dispatched} to devices that host the experts they are routed to via an all-to-all communication (All-to-All) \citep{shoeybi2019megatron}. Then, each device performs the feed-forward operation with its local expert weights. The tokens' output are then routed, or \textit{combined}, back to their original devices with another All-to-All (see \cref{fig:method}).

EP is designed with an assumption that load per GPU is always approximately balanced. However, in practice, that is rarely the case. Well-trained MoE models are shown to exhibit consistent imbalanced expert routing \citep{deepseek_v3_liu2024,gptoss_agarwal2025}, and arguably for a good reason -- as MoE layer training converges, some experts may become specialized to a certain domain or task, while others become generalized across a broad range of knowledge. During domain- or task-specific post-training or inference, only experts relevant to the tasks at hand are mostly activated while others stay dormant. So imbalanced routing, except for expert collapse, is a natural and desirable behavior for MoE models \citep{specialized_moe_qiu2025demons}. During post-training or inference, parameter-altering load balancing, like auxiliary losses, is discouraged or not allowed to preserve the integrity of the pre-trained MoE routing behaviors \citep{huang2024toward,locality_aware_hu2025communication}.
Standard EP, as such, is not designed to handle this phenomenon efficiently. Under worst-case imbalanced scenarios, EP may concentrate an overwhelming number of tokens from all devices to a few overloaded devices. This may cause high computational latency or even out-of-memory (OOM) failures if a GPU cannot store or process the excess tokens. Naive mitigation methods, such as lowering the batch size reduce throughput and increase latency. Advanced strategies like using redundant experts \citep{deepseek_v3_liu2024}, meanwhile, increase memory consumption, is only applicable for inference and still fails in the  worst cases.

To tackle this problem, we propose \textbf{Least-Loaded Expert Parallelism (\ourmethod)}, a novel EP algorithm that dynamically routes excess tokens, along with their corresponding expert weights, from overloaded devices to underloaded devices. Conceptually, when the globally assigned load on an expert group on a GPU exceeds a certain threshold, \ourmethod will only assign tokens to that GPU up to the capacity threshold, and then transfer the remaining tokens as well as the corresponding expert weights to the least-loaded devices for them to ``share'' the excess workload.
\ourmethod is designed to distributed workloads and memory usage across devices such that all devices complete their tasks roughly at the same time with minimal latency, while maintaining minimal peak memory usage per GPU.
Our load balancing algorithm is also designed to be conscious not only about compute load, but also about per-GPU memory allocations and cross-GPU communication overhead. Specifically, an excess tokens transfer is only triggered when the cost of transferring the tokens is less than the cost of processing them locally. Moreover, \ourmethod comes with backward-pass support, which allows it to be applied dynamically at every iteration of the training loop as well as inference. Importantly, LLEP is an \textbf{exact} MoE computation algorithm. Unlike other methods \citep{specialized_moe_qiu2025demons,locality_aware_hu2025communication}, it does not alter the models' behaviors for the sake of efficiency.

\ourmethod demonstrates significant speedup and peak-memory reduction over standard EP. As shown in \Figref{fig:speedup}, it achieves up to 4.6$\times$ speedup under extremely imbalanced scenarios, while maintaining a similar throughput as standard EP when the routing is balanced. Regarding peak memory usage per GPU, \ourmethod maintains a relatively stable peak-memory consumption across all scenarios, while standard EP's memory usage grows dramatically with imbalance, up to 4$\times$, which will cause OOM crashes if any GPU does not have enough reserve. As such, \ourmethod is able to handle large models with fewer GPUs while maintaining high throughput. In end-to-end testing with full pre-trained models, LLEP achieves up to 1.4$\times$ and 1.9$\times$ speedups for gpt-oss-20b and gpt-oss-120b respectively. We conduct a comprehensive theoretical and empirical analysis to provide insights into the cost dynamics of MoEs and expert parallelism, and discuss the trade-offs and design knobs for hardware-specific configuration tuning for best performance.




\section{Background}\label{sec:background}


\subsection{Mixture-of-Experts}\label{sec:background:moe}



Since the pioneering work of \citet{lepikhin2020gshard}, MoE models have become the de-facto standard for horizontally scaling LLMs, with DeepSeek-V3 \citep{deepseek_v3_liu2024}, gpt-oss \citep{gptoss_agarwal2025}, Kimi-K2 \citep{kimi_k2_team2025kimi} as notable examples.
The MoE feed-forward layer enables models to learn extensive knowledge across different expert weight matrices, while allowing individually tokens to be processed by a sparse subset of experts, thus improving efficiency and scalability. 
While concrete implementations vary, MoE architectures rely on the core idea of a \textit{router} layer, which selects the top-$K$ experts to route each token. Tokens are subsequently processed by each activated expert, which is typically constructed as a feed-forward (FFN) layer.
Formally, given a token $x$'s hidden representation $\vu \in \mathbb{R}^{D}$, the MoE module has $N$ experts $\{\text{FFN}_0, \text{FFN}_1, \ldots, \text{FFN}_{N-1}\}$ and a router layer $\text{Router}$ that selects the top-$K$ experts to route $x$ to. For brevity, we define $\text{FFN}_i(\vu) = \vu^T \mW_i$ where $\mW_i \in \mathbb{R}^{D \times H}$ is the weight matrix of expert $i$, and $\mW_r \in \mathbb{R}^{D \times N}$ is the weight matrix of the router layer. Then, the MoE output $\vh$ is
\begin{align}\label{eqn:moe_output}
    \vh = \sum_{i=0}^{N-1} g_{i} \text{FFN}_i(\vu), \quad \text{where} \quad
    g_{i} &= 
    \begin{cases}
        s_i, & \text{if } s_i \in \text{top-}K(\{s_j \mid 0 \leq j \leq N-1\}, K) \\
        0,   & \text{otherwise}
    \end{cases}\\
    s_i &= \text{softmax}_i (\vu^T \mW_r).
\end{align}
Above, $g_{i}$ is the gating affinity score for expert $i$, and only top-$k$ highest-scoring experts are selected to contribute to the output.


\paragraph{Implementation.} The number of GEneral Matrix Multiplications (GEMMs) required to process one token scales with the number of activated experts. Naive implementations (e.g., looping through each expert sequentially per token) can be extremely inefficient. A simple approach to improving performance is efficiently forming per-expert token batches via re-indexing. Assume that for a batch of tokens $\mB \in \mathbb{R}^{B \times D}$, $G \leq N$ experts are activated, i.e., receive routed tokens. Without loss of generality, assume these $G$ activated experts are experts $0, \ldots, G-1$. Then, we can form a re-indexed $\mB^\prime = [\mB_0, \mB_1,\ldots, \mB_{G-1}]$, where $\mB_i \in \mathbb{R}^{B_i \times D}$ consists of all tokens routed to expert $i$. As an example, if $\mB$ consists of four tokens $\mB = [\va, \vb, \vc, \vd]$ routed to experts $[2, 6, 2, 1]$, then we can form $\mB^\prime = [\vd, \va, \vc, \vb]$ with $\mB_1 = [\vd], \mB_2 = [\va, \vc]$, and $\mB_6 = [\vb]$. Under this re-indexing approach, each MoE layer computes $G$ GEMM operations $\mB_i\mW_i$ for experts $i = 0, 1,\ldots, G-1$.




\begin{algorithm}[tb]
  \caption{Highly Efficient Expert Parallelism dispatch\_combine: operation per device $p$ (zero-indexed), EP world size $P$, number of experts per device $M=N/P$}
  \label{alg:ep}
  \begin{algorithmic}
    \STATE {\bfseries Input:} $K$-repeated input tokens $\mathcal{B}_p \in \mathbb{R}^{B_p \times K \times D}$, router weights $\mathcal{G}_p \in \mathbb{R}^{B_p \times K}$, router indices $\mathcal{I}_p \in \mathbb{Z}^{B_p \times K}$, local expert weights $\mW_i \in \mathbb{R}^{D \times H}$ for $i=pM,pM + 1,\ldots,(p+1)M - 1$
    \STATE \textcolor{gray}{// Perform dispatch: route tokens and weights to devices that host the experts they are routed to}
    \STATE $\bar{\mathcal{I}_p}$ $\leftarrow$ sort(flatten($\mathcal{I}_p$, dims=[$B_p$, $K$]))
    \STATE $\bar{\mathcal{B}_p}$ $\leftarrow$ index\_select(flatten($\mathcal{B}_p$, dims=[$B_p$, $K$]), $\bar{\mathcal{I}_p}$)
    \STATE $\bar{\mathcal{G}_p}$ $\leftarrow$ index\_select(flatten($\mathcal{G}_p$, dims=[$B_p$, $K$]), $\bar{\mathcal{I}_p}$)
    \STATE \{$\mB_i | i \in \text{unique}(\bar{\mathcal{I}_p})$\} $\leftarrow$ slice($\bar{\mathcal{B}_p}$)
    \STATE \{$\mG_i | i \in \text{unique}(\bar{\mathcal{I}_p})$\} $\leftarrow$ slice($\bar{\mathcal{G}_p}$)
    \STATE \{$\hat{\mB_i} | i \in [pM,(p+1)M - 1]$\} $\leftarrow$ All-to-All(\{$\mB_i$\})
    \STATE \{$\hat{\mG_i} | i \in [pM,(p+1)M - 1]$\} $\leftarrow$ All-to-All(\{$\mG_i$\})
    \STATE \textcolor{gray}{// compute Grouped-GEMMs for local experts}
    \STATE \{$\hat{\mH_i} = \hat{\mG_i} \odot \hat{\mB_i}\mW_i | i \in [pM,(p+1)M) - 1]$\}
    \STATE \textcolor{gray}{// Perform combine: route outputs to where they originally came from}
    \STATE \{$\mH_i$\} $\leftarrow$ All-to-All-reverse(\{$\hat{\mH_i}$\})
    \STATE \textcolor{gray}{// reverse the sorting and reindexing}
    \STATE $\bar{\mathcal{H}_p}$ $\leftarrow$ concat(\{$\mH_i$\})
    \STATE $\mathcal{H}_p$ $\leftarrow$ reverse\_sort($\bar{\mathcal{H}_p}$, $\bar{\mathcal{I}_p}$)de
    \STATE $\mathcal{H}_p$ $\leftarrow$ reshape($\mathcal{H}_p$, $(B_p, K, H)$)
    \STATE $\mathcal{H}_p'$ $\leftarrow$ sum($\mathcal{H}_p$, dim=K)
    \STATE {\bfseries Output:} $\mathcal{H}_p'$  
  \end{algorithmic}
\end{algorithm}

\subsection{Expert Parallelism}\label{sec:background:ep}



To train large MoE models across multiple GPUs, expert parallelism (EP) is generally preferred over tensor or pipeline parallelism \citep{shoeybi2019megatron}, as it enables more efficient utilization of memory and communication bandwidth. In EP, experts are distributed across GPUs, with each device hosting only a local subset of experts.
During computation, input tokens are first processed by a router layer to produce global routing indices and corresponding routing weights (affinity scores). The resulting (input, indices, weight) tuples are then sorted and re-indexed, and subsequently dispatched to the GPUs that host the selected experts, as determined by the routing indices.

The routing process is typically conducted using the \emph{dispatch-combine} procedure. \Cref{alg:ep} formalizes a highly-efficient per-device implementation of this procedure, while \Cref{fig:standard_ep} provides a visual illustration of EP under an imbalanced routing scenario. Specifically, during \textit{dispatch}, each device sends its local input tokens to foreign experts' devices and receives foreign tokens assigned to its local experts. This data exchange paradigm is called an \emph{All-to-All} communication operation \citep{bruck_all2all_sewell2024bruck,optim_all2all_punniyamurthy2024optimizing}. After expert FFN computation is completed, outputs are aggregated in the \textit{combine} stage. Here, all expert outputs are sent back to their originating devices via another All-to-All operation. Beyond standard \href{https://docs.nvidia.com/deeplearning/nccl/user-guide/docs/index.html}{NCCL}-based collectives, EP can be implemented more efficiently at the kernel level using specialized libraries such as \href{https://github.com/deepseek-ai/DeepEP}{DeepEP} \citep{deepseek_v3_liu2024} and \href{https://github.com/ByteDance-Seed/Triton-distributed}{Triton-Distributed} \citep{zheng2025tritondistributed}.

\section{Analysis}\label{sec:background:analysis}
\subsection{Imbalanced Routing}\label{sec:background:imbalanced_routing}


\begin{figure}[t]
\centering
\begin{subfigure}[b]{0.48\columnwidth}
    \centering
    \includegraphics[width=\textwidth]{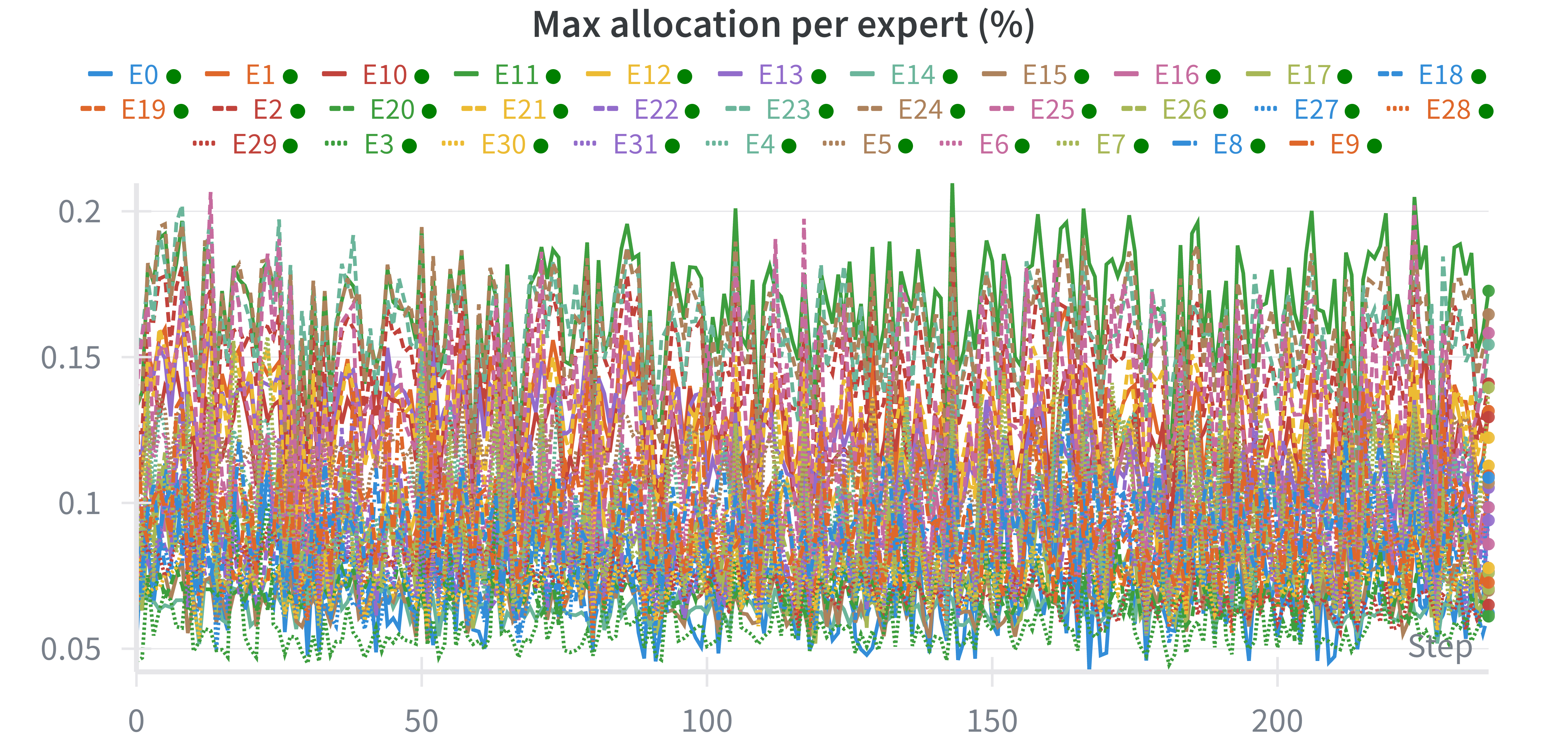}
    \caption{Max load per expert.}
    \label{fig:gptoss20b_expert_alloc}
\end{subfigure}
\hfill
\begin{subfigure}[b]{0.48\columnwidth}
    \centering
    \includegraphics[width=\textwidth]{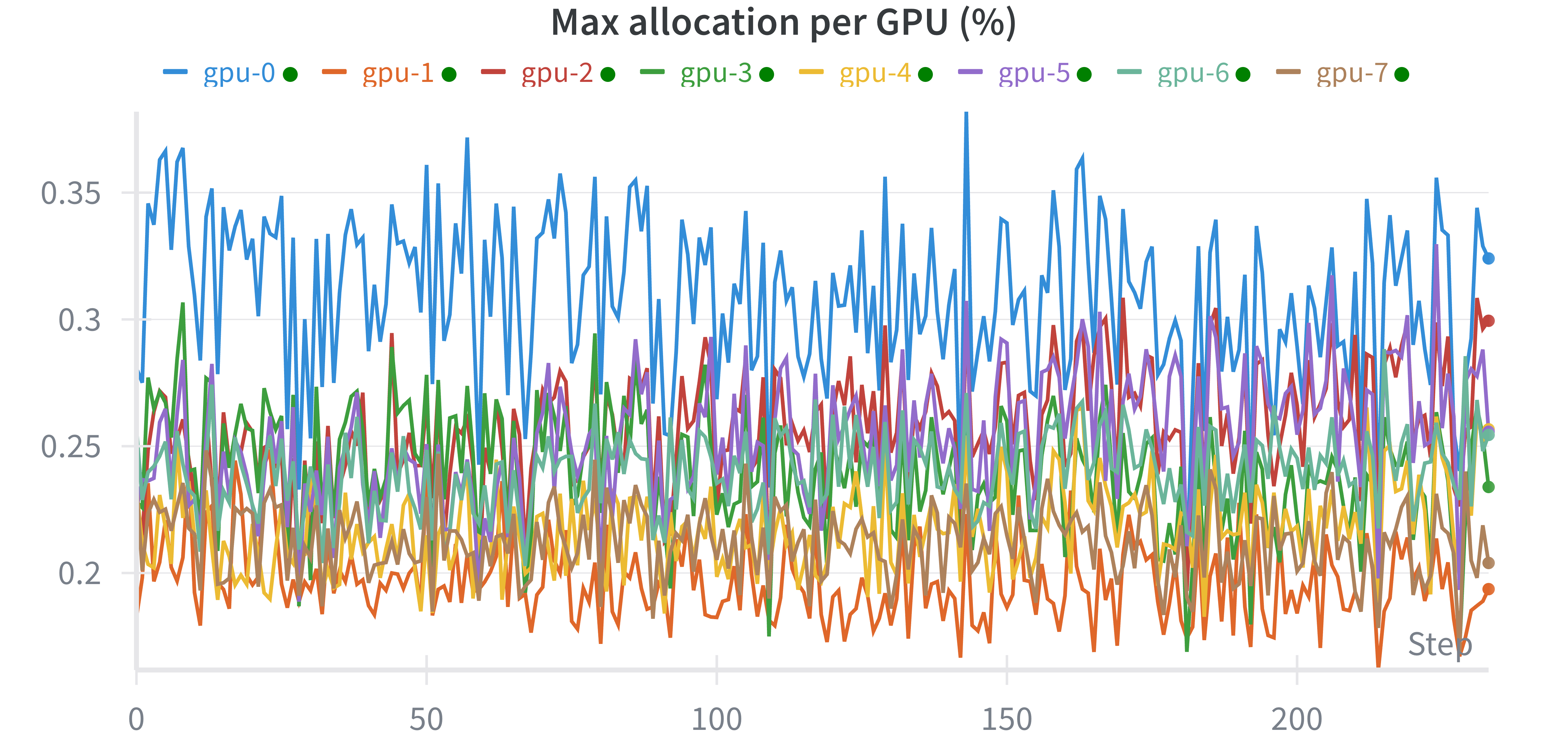}
    \caption{Max load per GPU (8-way EP).}
    \label{fig:gptoss20b_gpu_alloc}
\end{subfigure}
\caption{Expert routing imbalances across all layers of gpt-oss-20b across batches of a math dataset. \textbf{(a)} E11 has up to 20\% load vs.\ $\sim$3\% balanced. \textbf{(b)} GPU 0 has 30-35\% vs.\ $\sim$12.5\% balanced. Note that the load numbers do not add up to 100\% because values are maximums across all layers. 
}
\label{fig:gptoss20b_alloc}
\end{figure}

%
%

Even large and high-performing MoE models~\citep{deepseek_v3_liu2024,lmsys2025ep} have been shown to experience imbalance expert routing, where tokens are routed to a small subset of experts. 
We investigate the specific dynamics of token-routing by running gpt-oss-20b~\citep{gptoss_agarwal2025}, which has 32 experts, through many batches of data, under 8-way EP (each GPU hosts 4 experts). To keep the data distribution familiar, we feed the model with conversation data where the questions come from DeepScaleR~\citep{deepscaler2025} and response chain-of-thoughts are generated from gpt-oss-20b itself.
In~\Cref{fig:gptoss20b_expert_alloc}, we observe that tokens are consistently routed to certain expert positions, with position E11 dominating. This means that at least one 11th expert of the 24 MoE layers consistently receives dominant load across data batches. Despite that, the GPU that E11 is located on, gpu-2, does not have the highest load; gpu-0, hosting experts E0-E3, has the highest load among all GPUs. This implies that certain GPU devices may handle an overwhelming number of tokens under extremely imbalanced routing. Interestingly, while E11 typically takes on the most tokens, certain batches result in more tokens routed to other experts. That is, the degree of imbalance changes on a per-batch basis.


\paragraph{Are imbalanced MoE models actually bad?} It may be tempting to attribute imbalanced MoE routing to pre-training deficiencies, like skewed data or poorly designed load-balancing losses, e.g., \citet{zhou2022mixture,shazeer2017outrageously}. It is true that if not carefully trained, MoE models can exhibit ``expert collapse'', where only a small subset of experts is ever activated, producing sub-optimal or weak models. However, as we showed previously, even state-of-the-art MoE models exhibit some degree of imbalanced routing, albeit to a much milder degree than extreme expert collapse.

Rather than aiming for perfectly balanced routing, we consider mild imbalance a natural property of a well-trained MoE model. After undergoing large-scale pre-training, subsets of experts often specialize in particular knowledge domains, tasks, or capabilities \citep{specialized_moe_qiu2025demons,locality_aware_hu2025communication,song2025mixture}. Consequently, when an MoE model is further fine-tuned or evaluated on a specific domain, such as mathematics, experts specialized for that domain are activated more frequently. This leads to imbalanced routing. At the same time, some experts may evolve into broadly applicable, domain-agnostic ``shared'' experts that consistently handle generic linguistic patterns, such as grammar, across tasks. This phenomenon has also been observed and embraced in prior work \citep{deepseek_v3_liu2024}. From this perspective, aggressively enforcing balanced routing, e.g., by altering model behavior through auxiliary load-balancing losses \citep{switch_transformers_fedus2022switch} or moving-average routing biases \citep{deepseek_v3_liu2024}, risks disrupting these learned specialization patterns within particular experts. Instead of correcting imbalance at the model level, we instead embrace it, proposing a \textit{system-level} mechanism for both training and inference for maximizing throughput under imbalanced routing. This respects the inherent specialization among experts while mitigating inefficiencies that arise with imbalance.

\begin{algorithm}[t]
  \caption{Least-Loaded Assignment (LLA): Calculate a plan of that assign an expert' portions of tokens to different devices, as well as a corresponding weight transfer plan.}
  \label{alg:least_loaded_assignment}
  \begin{algorithmic}
    \STATE {\bfseries Input:} global expert loads $\vl \in \mathbb{R}^{N}$, \# local experts $M$, factor $\alpha$, minimum tokens per GEMM $m$
    \STATE \textcolor{gray}{// $\hat{\vl}$ is sorted loads, $I_{\hat{\vl}}$ is sorted indices}
    \STATE $\hat{\vl},I_{\hat{\vl}}$ $\leftarrow$ sort($\vl$, decreasing=true)
    \STATE \textcolor{gray}{// native/pending/assigned load per GPU}
    \STATE $g_n \in \mathbb{Z}^{P}$ $\leftarrow$ sum of loads of local experts
    \STATE $g_p$ $\leftarrow$ $g_n$
    \STATE $g_a \in \mathbb{Z}^{P}$ $\leftarrow$ 0 (zeros)
    \STATE \textcolor{gray}{// max tokens per GPU allowed}
    \STATE $m_{\alpha}$ $\leftarrow$ $\alpha \times \frac{1}{P} \times \sum_{i=0}^{N-1} \hat{\vl}_i$
    \STATE $\mathcal{A}$ $\leftarrow$ \{\} \textcolor{gray}{// assignments map for each expert}
    \FOR{$i$, $e$ in zip($I_{\hat{\vl}},\hat{\vl}$)}
      \STATE $ng$ $\leftarrow$ $\text{floor}(i / M)$
      \STATE $g_p[ng]$ $\leftarrow$ $g_p[ng] - e$
      \STATE \textcolor{gray}{// available tokens on native GPU}
      \STATE $na$ $\leftarrow$ $m_{\alpha} - g_a[ng] - g_p[ng]$
      \STATE $A$ $\leftarrow$ [] \textcolor{gray}{// assignments}
      \IF{$na \geq e$}
        \STATE \textcolor{gray}{// Case 1: Native GPU can handle all tokens}
        \STATE $A$ $\leftarrow$ $A$ + [(ng, 0, e)]
        \STATE $g_a[ng]$ $\leftarrow$ $g_a[ng] + e$
      \ELSIF{$na > 0$}
        \STATE \textcolor{gray}{// Case 2: Native GPU takes what it can, spill the rest to other GPUs}
        \STATE $nc$ $\leftarrow$ min($na$, $e$)
        \STATE $to$ $\leftarrow$ $nc$ \textcolor{gray}{// token offset}
        \STATE $A$ $\leftarrow$ $A$ + [(ng, 0, nc)]
        \STATE $g_a[ng]$ $\leftarrow$ $g_a[ng] + nc$
        \STATE $r$ $\leftarrow$ $e - nc$ \textcolor{gray}{// remaining}
        \STATE Call LLAS($ng$, $r$, $to$, $A$, $g_a$, $g_p$, $m_{\alpha}$, $m$) in \cref{alg:least_loaded_assignment_spill}
      \ELSE
        \STATE \textcolor{gray}{// Case 3: Native GPU overflowed, spill entire expert work to other GPUs}
        \STATE Call LLAS($ng$, $e$, 0, $A$, $g_a$, $g_p$, $m_{\alpha}$, $m$)
      \ENDIF
      \STATE $\mathcal{A}[i]$ $\leftarrow$ $A$
    \ENDFOR
    \STATE $\mathcal{W}$ $\leftarrow$ construct weight transfer plan from $\mathcal{A}$
    \STATE {\bfseries Output:} $\mathcal{A}$, $\mathcal{W}$
  \end{algorithmic}
\end{algorithm}

Several approaches have been proposed to mitigate routing imbalance under EP. A naive solution is to reduce the batch size, but this severely degrades throughput. Another strategy employs chained gradient checkpointing to process tokens in smaller chunks; however, this approach remains inefficient and is still constrained by a hard memory ceiling. For inference, \citet{deepseek_v3_liu2024} propose an EP Load Balancer (EPLB) that replicates heavily loaded experts across devices based on time-delayed routing statistics. While effective in some settings, this method incurs additional memory overhead, is not applicable to fine-tuning; further, this can still result in out-of-memory (OOM) failures under extreme routing imbalance. In contrast, \citet{huang2024toward} suggests reserving additional memory for excess experts, which likewise incurs CPU and GPU memory overhead. 

\subsection{Distributed Latency and Memory Analysis}\label{sec:background:cost_analysis}

To gain deeper insight into the worst-case cost model of MoE layers under EP, we analyze both latency and peak memory usage in a holistic manner. We first consider the computation local to a single GPU device. Given a batch routed to $G$ local experts, the MoE layer performs 
$G$ GEMM operations, and the total latency can be approximated as
\begin{equation}
    T_{\text{local}} = \sum_{i=0}^{G-1} (T_{\text{overhead}} + B_i \times T_{B_i,D,H})
\end{equation}
where $T_{\text{overhead}}$ denotes the kernel launch latency, and $T_{B_i,D,H}$ 
is the per-token compute time, which depends on the token count $B_i$ and model dimensions $D$ and $H$ (defined in \cref{sec:background:moe}). The efficiency of $T_{B_i,D,H}$ is directly impacted by how GEMM kernels are implemented, optimized, and tuned with respect to different input and output sizes and configurations. In general, GEMMs become more efficient as $B_i$, $D$ and $H$ increase. 
For example, with $D$ and $H$ fixed, $T_{B_1,D,H} < T_{B_2,D,H}$ when $B_1 > B_2$. Therefore, given a fixed number of FLOPs, executing a small number of large GEMMs is significantly more efficient than executing many small GEMMs. EP exploits this property by aggregating tokens across devices, thereby reducing the number of local experts $G$ and increasing the effective batch size $B_i$ per expert. \shafiq{done up to this.}


The many $T_{\text{overhead}}$ can be reduced to only one by using a fused grouped-GEMM kernel, but that is not always faster because singular hardware-optimized GEMM kernels (cuBLAS) are more efficient at large $D$, and $H$. \Cref{fig:group-gemm-benchmark} shows that, even though we compute the exact same FLOPs, the elapsed time increases with the number of experts, and launching many small \href{https://developer.nvidia.com/cublas}{cuBLAS} GEMMs is still faster than a single fused \href{https://triton-lang.org/main/getting-started/tutorials/08-grouped-gemm.html}{Triton grouped-GEMM kernel}\footnote{cuBLAS is proprietary software by NVIDIA that is highly optimized for the hardware, while the Triton grouped-GEMM is an agnostic implementation.}.
The peak memory usage of the MoE layer is defined approximately as:
\begin{equation}
  M_{\text{local}} = \sum_{i=0}^{G-1} (B_i \times D + D \times H + B_i \times H)
\end{equation}





Under standard expert parallelism, $B_i$ is the total number of tokens routed to expert $i$ from across all EP devices. In the worst case, $B_i$ may approach the global batch size, causing all tokens to be concentrated on a single device while others are idle. This causes spiking latency and memory usage, or even out-of-memory crashes for the overloaded device.

\Cref{fig:speedup,fig:memory} show the slowdown and peak memory usage of a standard EP setup under different imbalance scenarios. As shown in \Cref{fig:speedup}, EP could be 4.6x slower when 95\% of tokens are routed to a single expert compared to the balanced baseline. As for peak memory usage, EP's peak memory usage per GPU may grow up to 4x, potentially causing OOM errors.

\begin{algorithm}[t]
  \caption{Least-Loaded Assignment Spill (LLAS): Spilling the remaining tokens to other GPUs}
  \label{alg:least_loaded_assignment_spill}
  \begin{algorithmic}
    \STATE {\bfseries Input:} native GPU $ng$, remaining tokens $r$, token offset $to$, assignments $A$, assigned load $g_a$, pending load $g_p$, $m_{\alpha}$, $m$
    \WHILE{$r > 0$}
      \STATE $\vo \in \mathbb{Z}^{P-1}$ $\leftarrow$ other GPUs $g\neq ng$ sorted by $g_a[g] + g_p[g]$
      \FOR{$o$ in $\vo$}
        \STATE $c$ $\leftarrow$ min($r$, $m_{\alpha} - g_a[o] - g_p[o]$)
        \IF{$c < m$ and $r > c$}
          \STATE skip \textcolor{gray}{// chunk too small}
        \ENDIF
        \STATE $A$ $\leftarrow$ $A$ + $[(o, to, to + c)]$ \textcolor{gray}{// assign load}
        \STATE $g_a[o]$ $\leftarrow$ $g_a[o] + c$
        \STATE $r$ $\leftarrow$ $r - c$
        \STATE $to$ $\leftarrow$ $to + c$ \textcolor{gray}{// increment token offset}
        \STATE break
      \ENDFOR
      \IF{none of $\vo$ assigned}
        \STATE \textcolor{gray}{// force assign the least loaded GPU}
        \STATE $o$ $\leftarrow$ $\vo[0]$
        \STATE $A$ $\leftarrow$ $A$ + $[(o, to, to + r)]$
        \STATE $g_a[o]$ $\leftarrow$ $g_a[o] + r$
        \STATE $r$ $\leftarrow$ 0
      \ENDIF
    \ENDWHILE
  \end{algorithmic}
\end{algorithm}

\section{Least-Loaded Expert Parallelism (LLEP)}\label{sec:method}

We explain in detail how our proposed \ourmethod works. Conceptually, our method will detect ahead of time the degree of imbalance of the global routing according to per-expert loads. If the imbalance is lower than a threshold $\lambda$, then we consider the routing as balanced and proceed to the standard EP procedure. Otherwise, we will execute the least-loaded assignment algorithm (\cref{alg:least_loaded_assignment}) to determine for each GPU device that it needs to compute GEMMs for which experts and with how much portions of the global tokens routed to them. If the GPU does not contain an assigned expert as resident, it will import the expert from its host GPU. The assignment takes into account the overhead cost of weight and data transfers, in comparison to the latency and memory cost of processing the tokens only for local experts. \Cref{alg:least_loaded_assignment,alg:least_loaded_assignment_spill,alg:llep} formally describe our method in detail.

\paragraph{Constraints.} Our method works by making routing decisions that are subject to the some constraints. First, factor $\alpha$ in \cref{alg:least_loaded_assignment} determines how much maximum token capacity a GPU can handle, which we defined as $m_{\alpha} = \alpha \sum_{i=0}^{N-1} \hat{\vl}_i / P$ tokens. $m_{\alpha}$ is not necessarily a physical memory limit, but rather a threshold that the GPU is considered overloaded. If a local expert load exceeds $m_{\alpha}$, it will spill the excess load to other GPUs. Second, $m$ is the minimum tokens per GEMM for it to be efficient. If a local expert load exceed the local GPU's occupied capacity, but the excess is less than $m$, we consider it's not worth it to spill and instead force the local GPU to compute it despite over-capacity (see \cref{sec:background:cost_analysis}). Third, imbalance ratio threshold $\lambda$ is used to determine whether the global loads are relatively balanced, in which case we switch back to standard EP. The reason is that our method employs a greedy least-loaded assignment (LLA) algorithm (\cref{alg:least_loaded_assignment}) that would produce the same routing plan as standard EP anyway, while causing a tiny time overhead. Without skipping this imbalance ratio check, our method is shown to be slightly slower than standard EP under perfectly balanced scenarios. The optimal values for $\alpha$, $m$, and $\lambda$ depend on $N$, $P$, $B_p$, $K$, $D$, $H$, the overall model size, and the physical system configuration. Thus, we recommend to tune these values for each use case.

\paragraph{Elaboration.} The least-loaded assignment (LLA) algorithm (\cref{alg:least_loaded_assignment}) determines, for each expert, which GPUs handle which portions of the global expert's load. First, it sorts the expert loads in decreasing order. Then, it determines the GPU allocations for each expert from largest-load to smallest-load ones. For each expert, it first determines if the native GPU (the one that hosts the expert's weights) can handle all the tokens of the expert. If it can, it assigns all the tokens to the native GPU. If it cannot, it spills the excess tokens to the least-loaded available GPU up to the capacity threshold. If there are still remaining excess tokens, it will continue this spilling loop (LLAS, \cref{alg:least_loaded_assignment_spill}) until all the tokens are assigned. Once the tokens routing plan is finalized, it will also construct the weight transfer plan accordingly. For example, if excess load of expert $i$ native to GPU $p$ is spilled to GPU $q$, then the weight transfer plan will include a weight transfer operation from $p\rightarrow q$ for $W_i$. The LLA algorithm ensures that each GPU prioritize computing most, if not all, of the its local experts' load first before accepting foreign experts' load. This is to minimize the number of weight transfers required.
The final \ourmethod algorithm (\cref{alg:llep}) will then execute the dispatch-compute-combine operations according to the routing plans obtained from LLA. Specifically, for each device, in addition to GEMM computation for native experts, \ourmethod will also compute the GEMMs for foreign experts that are assigned to the device.
Unlike others, \ourmethod supports proper gradient propagation. During the backward pass, the gradients for the spilled expert weights are returned to their native devices and accumulated with their native gradients respectively.

\begin{algorithm}[t]
  \caption{LLEP dispatch\_combine: operation per device $p$ (zero-indexed), EP world size $P$, number of experts per device $M=N/P$}
  \label{alg:llep}
  \begin{algorithmic}
    \STATE {\bfseries Input:} $\mathcal{B}_p \in \mathbb{R}^{B_p \times K \times D}$, $\mathcal{G}_p \in \mathbb{R}^{B_p \times K}$, $\mathcal{I}_p \in \mathbb{Z}^{B_p \times K}$, $\mW_i \in \mathbb{R}^{D \times H}$ for $i=pM,pM + 1,\ldots,(p+1)M - 1$
    \STATE $\vl$ $\leftarrow$ sum of loads of global experts across all GPUs
    \IF{$\max(\vl) / \text{mean}(\vl) < \lambda$}
      \STATE Call standard EP \cref{alg:ep}
      \STATE {\bfseries Output:} $\mathcal{H}_p'$ from standard EP
    \ENDIF
    \STATE $\bar{\mathcal{I}_p}$ $\leftarrow$ sort(flatten($\mathcal{I}_p$, dims=[$B_p$, $K$]))
    \STATE $\bar{\mathcal{B}_p}$ $\leftarrow$ index\_select(flatten($\mathcal{B}_p$, dims=[$B_p$, $K$]), $\bar{\mathcal{I}_p}$)
    \STATE $\bar{\mathcal{G}_p}$ $\leftarrow$ index\_select(flatten($\mathcal{G}_p$, dims=[$B_p$, $K$]), $\bar{\mathcal{I}_p}$)
    \STATE \textcolor{gray}{// construct routing plan and weight transfer plan}
    \STATE $\mathcal{A}$, $\mathcal{W}$ $\leftarrow$ LLA($\vl$, $M$)
    \STATE \{$\mB_i | i \in [0,...,P]$\} $\leftarrow$ build chunks of $\bar{\mathcal{B}_p}$ from $\mathcal{A}$
    \STATE \{$\mG_i | i \in [0,...,P]$\} $\leftarrow$ build chunks of $\bar{\mathcal{G}_p}$ from $\mathcal{A}$
    \STATE \textcolor{gray}{// S is expert IDs of foreign experts assigned to this device}
    \STATE \{$\hat{\mB_i} | i \in [pM,(p+1)M - 1] \cup S$\} $\leftarrow$ All-to-All(\{$\mB_i$\})
    \STATE \{$\hat{\mG_i} | i \in [pM,(p+1)M - 1] \cup S$\} $\leftarrow$ All-to-All(\{$\mG_i$\})
    \STATE \{$\mW_j | j \in S$\} $\leftarrow$ P2P Transfer weights from other GPUs to this GPU
    \STATE \textcolor{gray}{// compute Grouped-GEMMs for native and foreign experts}
    \STATE \{$\hat{\mH_i} = \hat{\mG_i} \odot \hat{\mB_i}\mW_i | i \in [pM,(p+1)M - 1] \cup S$\}
    \STATE \textcolor{gray}{// Perform combine: route outputs to where they originally came from}
    \STATE \{$\mH_i$\} $\leftarrow$ All-to-All-reverse(\{$\hat{\mH_i}$\})
    \STATE \textcolor{gray}{// reverse the sorting and reindexing}
    \STATE $\bar{\mathcal{H}_p}$ $\leftarrow$ concat(\{$\mH_i$\})
    \STATE $\mathcal{H}_p$ $\leftarrow$ reverse\_sort($\bar{\mathcal{H}_p}$, $\bar{\mathcal{I}_p}$)
    \STATE $\mathcal{H}_p$ $\leftarrow$ reshape($\mathcal{H}_p$, $(B_p, K, H)$)
    \STATE $\mathcal{H}_p'$ $\leftarrow$ sum($\mathcal{H}_p$, dim=K)
    \STATE {\bfseries Output:} $\mathcal{H}_p'$
  \end{algorithmic}
\end{algorithm}




\paragraph{Implementation \& Optimization.} In the experiments, we implement our method with the standard Torch's NCCL for All-to-All and peer-to-peer (P2P) operatives. The LLA algorithm is implemented in pure Python. While our simple LLEP implementation is already showing significant speedup and memory saving, there are further opportunities to optimize and reduce overhead. For instance, the communication operatives can be written as low-level C++/Triton kernels, or using a modified version of DeepEP \citep{deepseek_v3_liu2024}. Such a fused operative may also perform direct All-to-All on unsorted tensors $\mathcal{B}_p$ and $\mathcal{G}_p$, avoiding the memory-intensive index\_select operation (\cref{alg:llep}). The communication can be overlapped with computation or hidden behind the Grouped-GEMM operation. For multi-node setups, we can further modify LLEP to prefer spilling work to intra-node devices to limit the higher inter-node communication overhead.






\section{Experiments}\label{sec:experiments}

We show the advantages of \ourmethod under two settings: Controlled experiments, where we precisely simulate imbalanced loads across multiple popular MoE configurations, and end-to-end training, where we train a MoE model with SFT, comparing against standard EP. We conclude with an ablation study, characterizing various hyperparameters in LLEP.

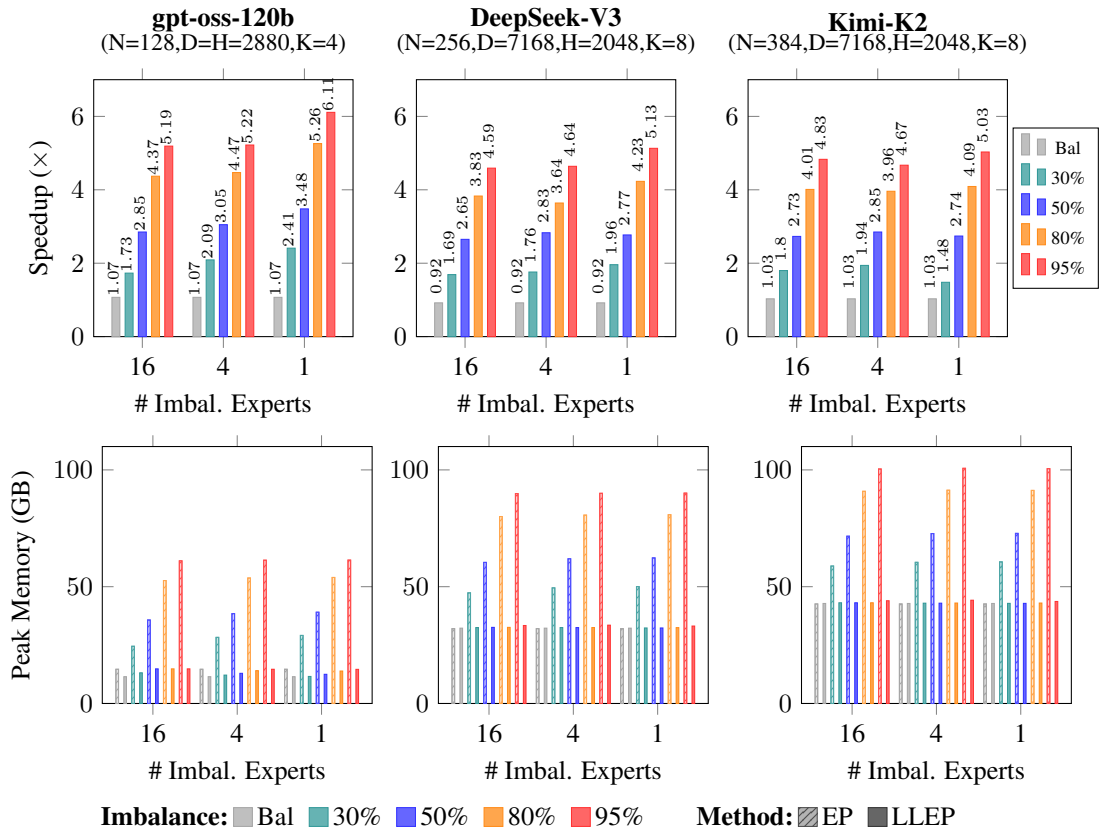
\begin{figure*}[t]
\centering
\begin{tabular}{@{}c@{\hskip 3pt}c@{\hskip 3pt}c@{}}
\begin{tikzpicture}
\begin{axis}[
    ybar,
    width=0.36\textwidth,
    height=5cm,
    ylabel={Speedup ($\times$)},
    symbolic x coords={16, 4, 1},
    xtick=data,
    xlabel={\# Imbal. Experts},
    title style={align=center}, 
    title={\textbf{gpt-oss-120b} \\[0.1em] {\footnotesize (N=128,D=H=2880,K=4)}}, 
    ymin=0,
    ymax=7,
    bar width=3pt,
    enlarge x limits=0.3,
    nodes near coords,
    nodes near coords style={font=\tiny, rotate=90, anchor=west},
    every node near coord/.append style={yshift=1pt,xshift=-3pt},
    clip=false,
]
\addplot[fill=gray!50, draw=gray!70] coordinates {(16, 1.07) (4, 1.07) (1, 1.07)};
\addplot[fill=teal!60, draw=teal!80] coordinates {(16, 1.73) (4, 2.09) (1, 2.41)};
\addplot[fill=blue!60, draw=blue!80] coordinates {(16, 2.85) (4, 3.05) (1, 3.48)};
\addplot[fill=orange!70, draw=orange!90] coordinates {(16, 4.37) (4, 4.47) (1, 5.26)};
\addplot[fill=red!60, draw=red!80] coordinates {(16, 5.19) (4, 5.22) (1, 6.11)};
\end{axis}
\end{tikzpicture}
&
\begin{tikzpicture}
\begin{axis}[
    ybar,
    width=0.36\textwidth,
    height=5cm,
    symbolic x coords={16, 4, 1},
    xtick=data,
    xlabel={\# Imbal. Experts},
    title style={align=center}, 
    title={\textbf{DeepSeek-V3} \\[0.1em] {\footnotesize (N=256,D=7168,H=2048,K=8)}}, 
    ymin=0,
    ymax=7,
    bar width=3pt,
    enlarge x limits=0.3,
    nodes near coords,
    nodes near coords style={font=\tiny, rotate=90, anchor=west},
    every node near coord/.append style={yshift=1pt},
    clip=false,
]
\addplot[fill=gray!50, draw=gray!70] coordinates {(16, 0.92) (4, 0.92) (1, 0.92)};
\addplot[fill=teal!60, draw=teal!80] coordinates {(16, 1.69) (4, 1.76) (1, 1.96)};
\addplot[fill=blue!60, draw=blue!80] coordinates {(16, 2.65) (4, 2.83) (1, 2.77)};
\addplot[fill=orange!70, draw=orange!90] coordinates {(16, 3.83) (4, 3.64) (1, 4.23)};
\addplot[fill=red!60, draw=red!80] coordinates {(16, 4.59) (4, 4.64) (1, 5.13)};
\end{axis}
\end{tikzpicture}
&
\begin{tikzpicture}
\begin{axis}[
    ybar,
    width=0.36\textwidth,
    height=5cm,
    symbolic x coords={16, 4, 1},
    xtick=data,
    xlabel={\# Imbal. Experts},
    title style={align=center}, 
    title={\textbf{Kimi-K2} \\[0.1em] {\footnotesize (N=384,D=7168,H=2048,K=8)}}, 
    ymin=0,
    ymax=7,
    legend style={at={(1.02,0.5)}, anchor=west, legend columns=1, font=\scriptsize},
    bar width=3pt,
    enlarge x limits=0.3,
    nodes near coords,
    nodes near coords style={font=\tiny, rotate=90, anchor=west},
    every node near coord/.append style={yshift=1pt},
    clip=false,
]
\addplot[fill=gray!50, draw=gray!70] coordinates {(16, 1.03) (4, 1.03) (1, 1.03)};
\addplot[fill=teal!60, draw=teal!80] coordinates {(16, 1.80) (4, 1.94) (1, 1.48)};
\addplot[fill=blue!60, draw=blue!80] coordinates {(16, 2.73) (4, 2.85) (1, 2.74)};
\addplot[fill=orange!70, draw=orange!90] coordinates {(16, 4.01) (4, 3.96) (1, 4.09)};
\addplot[fill=red!60, draw=red!80] coordinates {(16, 4.83) (4, 4.67) (1, 5.03)};
\legend{Bal, 30\%, 50\%, 80\%, 95\%}
\end{axis}
\end{tikzpicture}
\\
\begin{tikzpicture}
\begin{axis}[
    ybar,
    width=0.37\textwidth,
    height=5cm,
    ylabel={Peak Memory (GB)},
    xtick={0,2,4},
    xticklabels={16, 4, 1},
    xlabel={\# Imbal. Experts},
    ymin=0,
    ymax=110,
    bar width=1pt,
    enlarge x limits=0.3,
    clip=false,
]
\addplot[fill=gray!20, draw=gray!70, postaction={pattern=north east lines, pattern color=gray!70}] 
    coordinates {(0, 14.64) (2, 14.64) (4, 14.64)};
\addplot[fill=gray!50, draw=gray!70] 
    coordinates {(0, 11.47) (2, 11.47) (4, 11.47)};
\addplot[fill=teal!20, draw=teal!70, postaction={pattern=north east lines, pattern color=teal!70}] 
    coordinates {(0, 24.49) (2, 28.30) (4, 29.13)};
\addplot[fill=teal!60, draw=teal!80] 
    coordinates {(0, 13.13) (2, 12.15) (4, 11.57)};
\addplot[fill=blue!20, draw=blue!70, postaction={pattern=north east lines, pattern color=blue!70}] 
    coordinates {(0, 35.74) (2, 38.46) (4, 39.07)};
\addplot[fill=blue!60, draw=blue!80] 
    coordinates {(0, 14.82) (2, 12.91) (4, 12.49)};
\addplot[fill=orange!20, draw=orange!70, postaction={pattern=north east lines, pattern color=orange!70}] 
    coordinates {(0, 52.63) (2, 53.72) (4, 53.95)};
\addplot[fill=orange!70, draw=orange!90] 
    coordinates {(0, 14.82) (2, 14.05) (4, 13.89)};
\addplot[fill=red!20, draw=red!70, postaction={pattern=north east lines, pattern color=red!70}] 
    coordinates {(0, 61.07) (2, 61.34) (4, 61.38)};
\addplot[fill=red!60, draw=red!80] 
    coordinates {(0, 14.82) (2, 14.65) (4, 14.61)};
\end{axis}
\end{tikzpicture}
&
\begin{tikzpicture}
\begin{axis}[
    ybar,
    width=0.37\textwidth,
    height=5cm,
    xtick={0,2,4},
    xticklabels={16, 4, 1},
    xlabel={\# Imbal. Experts},
    ymin=0,
    ymax=110,
    bar width=1pt,
    enlarge x limits=0.3,
    clip=false,
]
\addplot[fill=gray!20, draw=gray!70, postaction={pattern=north east lines, pattern color=gray!70}] 
    coordinates {(0, 32.03) (2, 32.03) (4, 32.03)};
\addplot[fill=gray!50, draw=gray!70] 
    coordinates {(0, 32.25) (2, 32.25) (4, 32.25)};
\addplot[fill=teal!20, draw=teal!70, postaction={pattern=north east lines, pattern color=teal!70}] 
    coordinates {(0, 47.35) (2, 49.49) (4, 50.03)};
\addplot[fill=teal!60, draw=teal!80] 
    coordinates {(0, 32.50) (2, 32.50) (4, 32.33)};
\addplot[fill=blue!20, draw=blue!70, postaction={pattern=north east lines, pattern color=blue!70}] 
    coordinates {(0, 60.40) (2, 61.94) (4, 62.32)};
\addplot[fill=blue!60, draw=blue!80] 
    coordinates {(0, 32.58) (2, 32.50) (4, 32.33)};
\addplot[fill=orange!20, draw=orange!70, postaction={pattern=north east lines, pattern color=orange!70}] 
    coordinates {(0, 80.04) (2, 80.64) (4, 80.81)};
\addplot[fill=orange!70, draw=orange!90] 
    coordinates {(0, 32.58) (2, 32.50) (4, 32.47)};
\addplot[fill=red!20, draw=red!70, postaction={pattern=north east lines, pattern color=red!70}] 
    coordinates {(0, 89.86) (2, 89.98) (4, 90.06)};
\addplot[fill=red!60, draw=red!80] 
    coordinates {(0, 33.38) (2, 33.53) (4, 33.15)};
\end{axis}
\end{tikzpicture}
&
\begin{tikzpicture}
\begin{axis}[
    ybar,
    width=0.37\textwidth,
    height=5cm,
    xtick={0,2,4},
    xticklabels={16, 4, 1},
    xlabel={\# Imbal. Experts},
    ymin=0,
    ymax=110,
    bar width=1pt,
    enlarge x limits=0.3,
    clip=false,
]
\addplot[fill=gray!20, draw=gray!70, postaction={pattern=north east lines, pattern color=gray!70}] 
    coordinates {(0, 42.60) (2, 42.60) (4, 42.60)};
\addplot[fill=gray!50, draw=gray!70] 
    coordinates {(0, 42.82) (2, 42.82) (4, 42.82)};
\addplot[fill=teal!20, draw=teal!70, postaction={pattern=north east lines, pattern color=teal!70}] 
    coordinates {(0, 58.84) (2, 60.43) (4, 60.62)};
\addplot[fill=teal!60, draw=teal!80] 
    coordinates {(0, 43.11) (2, 42.94) (4, 42.86)};
\addplot[fill=blue!20, draw=blue!70, postaction={pattern=north east lines, pattern color=blue!70}] 
    coordinates {(0, 71.61) (2, 72.74) (4, 72.82)};
\addplot[fill=blue!60, draw=blue!80] 
    coordinates {(0, 43.11) (2, 42.94) (4, 42.86)};
\addplot[fill=orange!20, draw=orange!70, postaction={pattern=north east lines, pattern color=orange!70}] 
    coordinates {(0, 90.84) (2, 91.30) (4, 91.22)};
\addplot[fill=orange!70, draw=orange!90] 
    coordinates {(0, 43.11) (2, 43.00) (4, 43.00)};
\addplot[fill=red!20, draw=red!70, postaction={pattern=north east lines, pattern color=red!70}] 
    coordinates {(0, 100.39) (2, 100.68) (4, 100.52)};
\addplot[fill=red!60, draw=red!80] 
    coordinates {(0, 43.98) (2, 44.21) (4, 43.67)};
\end{axis}
\end{tikzpicture}
\\
\end{tabular}

\begin{tabular}{@{}r@{\,}l@{\;\;}l@{\;\;}l@{\;\;}l@{\;\;}l@{\quad\quad}r@{\,}l@{\;\;}l@{}}
\textbf{Imbalance:} & 
\tikz[baseline=-0.5ex]{\fill[gray!50, draw=gray!70] (0,-0.1) rectangle (0.25,0.15);}\,Bal &
\tikz[baseline=-0.5ex]{\fill[teal!60, draw=teal!80] (0,-0.1) rectangle (0.25,0.15);}\,30\% &
\tikz[baseline=-0.5ex]{\fill[blue!60, draw=blue!80] (0,-0.1) rectangle (0.25,0.15);}\,50\% &
\tikz[baseline=-0.5ex]{\fill[orange!70, draw=orange!90] (0,-0.1) rectangle (0.25,0.15);}\,80\% &
\tikz[baseline=-0.5ex]{\fill[red!60, draw=red!80] (0,-0.1) rectangle (0.25,0.15);}\,95\% &
\textbf{Method:} & 
\tikz[baseline=-0.5ex]{\fill[black!30, draw=black!70, postaction={pattern=north east lines, pattern color=black!70}] (0,-0.1) rectangle (0.25,0.15);}\,EP &
\tikz[baseline=-0.5ex]{\fill[black!60, draw=black!80] (0,-0.1) rectangle (0.25,0.15);}\,\ourmethod \\
\end{tabular}

\caption{Performance comparison of \ourmethod vs.\ standard EP across three MoE architectures. \textbf{Top row:} Speedup ($\times$, higher is better) of \ourmethod over standard EP. Gray bars show the balanced baseline ($\approx$1$\times$). Colored bars indicate imbalance levels (percentage of tokens routed to that many experts). Higher concentration yields greater speedup, up to 6.1$\times$ for GPT-OSS-120B. \textbf{Bottom row:} Peak memory usage per GPU (GB, lower is better). Hatched bars represent standard EP; solid bars represent \ourmethod. EP memory grows dramatically with imbalance (up to 100GB for Kimi-K2), while \ourmethod maintains near-constant memory across all scenarios.}
\label{fig:model_comparison}
\end{figure*}

\subsection{Speed and Memory Profiles}\label{sec:exp:speed_mem}


We analyze the speed and memory profiles of different popular MoE layers of different sizes and configurations, namely those used in gpt-oss-120b \citep{gptoss_agarwal2025}, DeepSeek-V3 \citep{deepseek_v3_liu2024} and Kimi-K2 \citep{kimi_k2_team2025kimi}. 
Unlike the simple case stated in \Cref{sec:background}, each MoE expert is a SwigGLU \citep{shazeer2020glu} feed-forward module that use three weight matrices instead of one.
We benchmark the forward pass speedups and peak memory consumption per GPU across $P = 8$ H200 GPUs. The batch size per GPU ($B$) 
is 32K for gpt-oss and 16K for DeepSeek-V3 and Kimi-K2. We simulate across different balanced and imbalanced routing scenarios, from 30\% to 95\% of tokens evenly concentrated into 1, 4 or 16 experts. For \ourmethod, we use $\lambda=1.3,\alpha=1,m=1024$.
\austin{How can we route to 1 expert? Is what it means to have x\% concentration for y experts explained anywhere?} 
\nxphi{each of the y expert got x/y\% uniformly, the other N-y experts each got (100-x)/(N-y). but I don't think we need to explain it clearly here}

\Cref{fig:model_comparison} summarizes the benchmarking results. As shown in the speedup row, across different configurations, \ourmethod outperforms standard EP across all imbalance scenarios, achieving greater speedup under more imbalanced routing, up to 6.11$\times$ for the most extreme case (95\% into 1 expert). Meanwhile, \ourmethod maintains EP's efficiency in perfectly balanced case, thanks to the adaptive ratio $\lambda$. In the peak memory row, \ourmethod maintains consistently and stably low memory consumption across all imbalance scenarios, with memory saving of up to 5$\times$, allowing us to increase the throughput (batch size) without running into out-of-memory (OOM) failure.

\subsection{End-to-End Full Model Speed Profiles In The Wild}\label{sec:exp:full_model_speed}

To measure the effectiveness of \ourmethod in the \textbf{wild}, instead of just simulating imbalances, we conduct end-to-end forward-pass throughput comparisons with real pre-trained gpt-oss-20b and gpt-oss-120b \citep{gptoss_agarwal2025} on samples from the Megatron-Math dataset \citep{du2025nemotron_math}, where the responses were generated by gpt-oss-120b itself. The results are reported in \Cref{fig:end_to_end_speedup}. Full model throughput is impacted by other irrelevant factors and fixed overheads, such as the attention layers. Thus, the speedup of the MoE layers is always greater than the reported numbers for the full model throughput. As shown, \ourmethod achieves up to 2.2$\times$ and 1.88$\times$ speedups for gpt-oss-20b and gpt-oss-120b respectively. Our method achieves better scaling efficiency with greater relative speedups the more GPUs are used. \ourmethod demonstrates its superiority because the pre-trained models exhibit imbalanced routing inherently even on in-domain data.

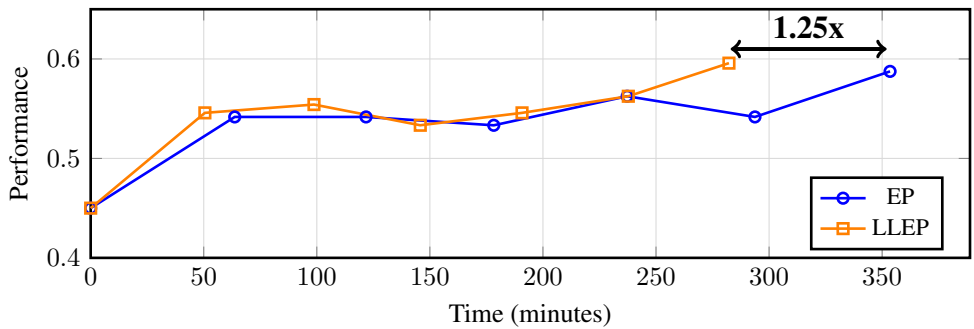
\begin{figure}[t]
\centering
\begin{tikzpicture}
  \begin{axis}[
    width=0.95\columnwidth,
    height=0.35\columnwidth,
    xlabel={Time (minutes)},
    ylabel={Performance},
    legend pos=south east,
    legend style={font=\small},
    grid=major,
    grid style={gray!30},
    line width=1pt,
    mark size=2pt,
    xmin=0,
    ymin=0.4,
    ymax=0.65,
  ]
  \addplot[color=blue, mark=o] coordinates {
    (0, 0.45)
    (63.7, 0.5417)
    (121.75, 0.5417)
    (178.25, 0.5333)
    (237.25, 0.5625)
    (293.75, 0.5417)
    (353.5, 0.5875)

  };
  \addlegendentry{EP}
    
  \addplot[color=orange, mark=square] coordinates {
    (0, 0.45)
    (50.5, 0.5458)
    (98.75, 0.55416)
    (145.75, 0.53333333)
    (190.75, 0.5458)
    (237.75, 0.5625)
    (282.25, 0.5958)

  };
  \addlegendentry{\ourmethod}
  \draw[<->, line width=1.5pt] (axis cs:283, 0.61) -- (axis cs:352, 0.61) node[midway, above, font=\large\bfseries] {1.25x};
  \end{axis}
\end{tikzpicture}
\caption{Performance (accuracy on AIME'25) vs.\ wall-time for EP and \ourmethod when training gpt-oss-20b (low effort) on full parameters using Zero-3 and CPU offloading for gradients and optimizer states. CPU operations and checkpoint saving introduce non-negotiable overheads. Despite additional overhead, training converges 1.25x faster with LLEP as compared to EP.}
\label{fig:ep_vs_llep_time_extended}
\end{figure}

\Cref{fig:ep_vs_llep_time_extended} shows the downstream performance (accuracy on AIME'25) vs.\ wall-clock time for EP and \ourmethod, when training gpt-oss-20b (low effort) on \textbf{full} parameters, using Zero-3 and CPU offloading for gradients and optimizer states. The training process requires more expensive and non-negotiable, but irrelevant, overheads, including on-CPU computations of gradients and parameter updates as well as checkpoint saving at each step.
As such, \ourmethod achieves 1.25$\times$ speedup over EP while achieving comparable accuracy.

\subsection{Ablation Study}\label{sec:exp:ablation}

We provide further insights into \ourmethod by ablating various factors and hyper-parameters.

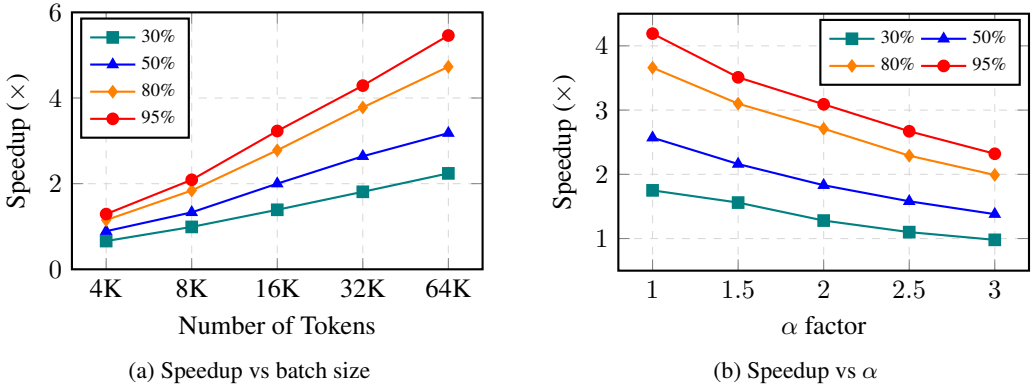
\begin{figure}[t]
\centering
\begin{subfigure}[b]{0.48\columnwidth}
\begin{tikzpicture}
\begin{axis}[
    width=1.05\textwidth,
    height=5cm,
    xlabel={Number of Tokens},
    ylabel={Speedup ($\times$)},
    xmode=log,
    log basis x=2,
    xtick={4096, 8192, 16384, 32768, 65536},
    xticklabels={4K, 8K, 16K, 32K, 64K},
    ymin=0,
    ymax=6,
    legend style={at={(0.02,0.98)}, anchor=north west, font=\scriptsize},
    grid=major,
    grid style={dashed, gray!30},
    mark size=2pt,
    line width=0.8pt,
]
\addplot[color=teal, mark=square*, thick] coordinates {
    (4096, 0.66) (8192, 0.99) (16384, 1.39) (32768, 1.81) (65536, 2.24)
};
\addplot[color=blue, mark=triangle*, thick] coordinates {
    (4096, 0.89) (8192, 1.33) (16384, 2.00) (32768, 2.64) (65536, 3.18)
};
\addplot[color=orange, mark=diamond*, thick] coordinates {
    (4096, 1.15) (8192, 1.84) (16384, 2.78) (32768, 3.78) (65536, 4.73)
};
\addplot[color=red, mark=*, thick] coordinates {
    (4096, 1.29) (8192, 2.09) (16384, 3.23) (32768, 4.29) (65536, 5.46)
};
\legend{30\%, 50\%, 80\%, 95\%}
\end{axis}
\end{tikzpicture}
\caption{Speedup vs batch size}
\label{fig:speedup_vs_tokens}
\end{subfigure}
\hfill
\begin{subfigure}[b]{0.48\columnwidth}
\begin{tikzpicture}
\begin{axis}[
    width=1.05\textwidth,
    height=5cm,
    xlabel={$\alpha$ factor},
    ylabel={Speedup ($\times$)},
    xtick={1.0, 1.5, 2.0, 2.5, 3.0},
    ymin=0.5,
    ymax=4.5,
    legend style={at={(0.98,0.98)}, anchor=north east, font=\scriptsize, legend columns=2},
    grid=major,
    grid style={dashed, gray!30},
    mark size=2pt,
    line width=0.8pt,
]
\addplot[color=teal, mark=square*, thick] coordinates {
    (1.0, 1.75) (1.5, 1.56) (2.0, 1.28) (2.5, 1.10) (3.0, 0.98)
};
\addplot[color=blue, mark=triangle*, thick] coordinates {
    (1.0, 2.57) (1.5, 2.16) (2.0, 1.83) (2.5, 1.58) (3.0, 1.38)
};
\addplot[color=orange, mark=diamond*, thick] coordinates {
    (1.0, 3.66) (1.5, 3.10) (2.0, 2.71) (2.5, 2.29) (3.0, 1.99)
};
\addplot[color=red, mark=*, thick] coordinates {
    (1.0, 4.19) (1.5, 3.51) (2.0, 3.09) (2.5, 2.67) (3.0, 2.32)
};
\legend{30\%, 50\%, 80\%, 95\%}
\end{axis}
\end{tikzpicture}
\caption{Speedup vs $\alpha$}
\label{fig:speedup_vs_alpha}
\end{subfigure}
\caption{Speedup of \ourmethod over standard EP with 4 imbalanced experts. (a) Speedup as a function of batch size. (b) Speedup as a function of $\alpha$; lower $\alpha$ yields higher speedups.}
\label{fig:ablation_speedup}
\end{figure}

\paragraph{Batch size $B$.} \Cref{fig:speedup_vs_tokens} shows that \ourmethod achieves greater speedups the more tokens we pack into the batch, across various imbalance scenarios. The reason is that large batch sizes saturate the capacity of each individual GPU and overwhelm any overhead introduced by the LLA algorithm (\cref{alg:least_loaded_assignment}), leading to a linear relationship between batch size and processing time. This means that the least collective processing time ($\max_i [\text{time-of-GPU } i]$) is achieved when compute workloads are evenly distributed across all GPUs.
The All-to-All data transfers of large batches also overshadow any associated weight transfer.

\paragraph{Factor $\alpha$} \Cref{fig:speedup_vs_alpha} shows the speedup curves across different $\alpha$ values. We observe that higher $\alpha$ leads the lower speedup, meaning allowing more per-GPU capacity than the balanced baseline before LLA spilling may result in inefficiency. This means that at large-enough batch sizes, \ourmethod prefers workloads to be balanced despite possibly higher communication costs.

%

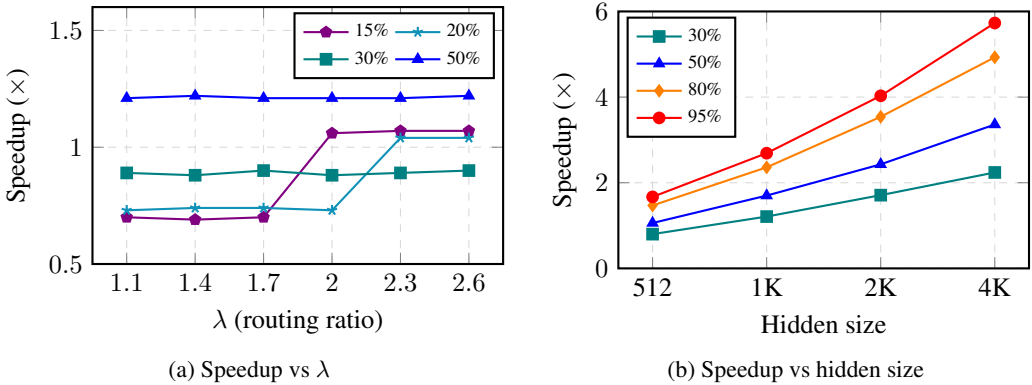
\begin{figure}[t]
\centering
\begin{subfigure}[b]{0.48\columnwidth}
\begin{tikzpicture}
\begin{axis}[
    width=1.05\textwidth,
    height=5cm,
    xlabel={$\lambda$ (routing ratio)},
    ylabel={Speedup ($\times$)},
    xtick={1.1, 1.4, 1.7, 2.0, 2.3, 2.6},
    ymin=0.5,
    ymax=1.6,
    legend style={at={(0.98,0.98)}, anchor=north east, font=\scriptsize, legend columns=2},
    grid=major,
    grid style={dashed, gray!30},
    mark size=2pt,
    line width=0.8pt,
]
\addplot[color=violet, mark=pentagon*, thick] coordinates {
    (1.1, 0.70) (1.4, 0.69) (1.7, 0.70) (2.0, 1.06) (2.3, 1.07) (2.6, 1.07)
};
\addplot[color=cyan!70!black, mark=star, thick] coordinates {
    (1.1, 0.73) (1.4, 0.74) (1.7, 0.74) (2.0, 0.73) (2.3, 1.04) (2.6, 1.04)
};
\addplot[color=teal, mark=square*, thick] coordinates {
    (1.1, 0.89) (1.4, 0.88) (1.7, 0.90) (2.0, 0.88) (2.3, 0.89) (2.6, 0.90)
};
\addplot[color=blue, mark=triangle*, thick] coordinates {
    (1.1, 1.21) (1.4, 1.22) (1.7, 1.21) (2.0, 1.21) (2.3, 1.21) (2.6, 1.22)
};

\legend{15\%, 20\%, 30\%, 50\%}
\end{axis}
\end{tikzpicture}
\caption{Speedup vs $\lambda$}
\label{fig:speedup_vs_lambda}
\end{subfigure}
\hfill
\begin{subfigure}[b]{0.48\columnwidth}
\begin{tikzpicture}
\begin{axis}[
    width=1.05\textwidth,
    height=5cm,
    xlabel={Hidden size},
    ylabel={Speedup ($\times$)},
    xmode=log,
    log basis x=2,
    xtick={512, 1024, 2048, 4096},
    xticklabels={512, 1K, 2K, 4K},
    ymin=0,
    ymax=6,
    legend style={at={(0.02,0.98)}, anchor=north west, font=\scriptsize},
    grid=major,
    grid style={dashed, gray!30},
    mark size=2pt,
    line width=0.8pt,
]
\addplot[color=teal, mark=square*, thick] coordinates {
    (512, 0.80) (1024, 1.21) (2048, 1.71) (4096, 2.24)
};
\addplot[color=blue, mark=triangle*, thick] coordinates {
    (512, 1.06) (1024, 1.70) (2048, 2.43) (4096, 3.36)
};
\addplot[color=orange, mark=diamond*, thick] coordinates {
    (512, 1.47) (1024, 2.36) (2048, 3.54) (4096, 4.93)
};
\addplot[color=red, mark=*, thick] coordinates {
    (512, 1.67) (1024, 2.69) (2048, 4.03) (4096, 5.73)
};
    
\legend{30\%, 50\%, 80\%, 95\%}
\end{axis}
\end{tikzpicture}
\caption{Speedup vs hidden size}
\label{fig:speedup_vs_hidden_size}
\end{subfigure}
\caption{Speedup of \ourmethod over standard EP with 4 imbalanced experts. (a) Speedup as a function of $\lambda$. (b) Speedup scales with hidden size.}
\label{fig:ablation_lambda_hidden}
\end{figure}

\paragraph{Adaptive ratio $\lambda$} \Cref{fig:speedup_vs_lambda} shows how the adaptive ratio $\lambda$ impacts speedup. Specifically, when the batch size is low ($B=8\text{K}$), we observe higher $\lambda$ is beneficial when the imbalance degree is low (15-20\%). In other words, it is better to revert to standard EP when the routing distribution is balanced enough that the overhead costs of \ourmethod's weight transfers surpassed the benefits of even computation.

\paragraph{Hidden size $D$ and $H$} \Cref{fig:speedup_vs_hidden_size} demonstrates that \ourmethod shines as we scale up the model's hidden sizes, despite the fact that its spilling weight transfers cost more.
The reason for this scaling effect is that as the hidden size increases, the compute efficiency of each GEMM improves compared to the data communication costs. In addition, similar to scaling batch sizes, large hidden sizes also saturate the GPU capacity. This causes the benefits of the compute workloads being perfectly balanced to overshadow any inconvenient weight transfer overhead.

\section{Conclusion}
We present least-loaded expert parallelism (\ourmethod), a novel EP algorithm that dynamically performs load balancing to address the MoE imbalanced routing phenomenon, while ensuring the exact MoE mathematical computation. \ourmethod achieves up to 5-6$\times$ speedups and 5$\times$ reduction in peak memory consumption for the MoE layers. It improves the end-to-end full-model throughputs of gpt-oss-120b by up to 90\%.

\bibliography{iclr2026_conference}
\bibliographystyle{iclr2026_conference}

\appendix
\newpage
\appendix
\onecolumn
\section{Additional Results}

\subsection{Separate vs Fused Grouped-GEMM}\label{sec:app:grouped_gemm}

\begin{figure}[h]
\centering
\begin{tikzpicture}
  \begin{axis}[
    width=0.5\columnwidth,
    height=0.4\columnwidth,
    xlabel={\# Experts (log-scale)},
    ylabel={Time (ms)},
    symbolic x coords={4,16,32,64,128,256},
    xtick=data,
    legend pos=north west,
    legend style={font=\small},
    grid=major,
    grid style={gray!30},
    line width=1pt,
    mark size=2pt,
  ]
  \addplot[color=blue, mark=o] coordinates {
    (4, 12.595264)
    (16, 12.275600)
    (32, 12.226976)
    (64, 12.513104)
    (128, 12.986288)
    (256, 14.880896)
  };
  \addlegendentry{cuBLAS}
  
  
  \addplot[color=orange, mark=triangle] coordinates {
    (4, 13.218112)
    (16, 13.644288)
    (32, 13.504576)
    (64, 14.536704)
    (128, 17.301537)
    (256, 18.902529)
  };
  \addlegendentry{Triton + TMA}
  \end{axis}
\end{tikzpicture}
\caption{Grouped-GEMM benchmark (\textbf{lower is better}): execution time vs.\ number of experts under balanced workload with the \textbf{same} total FLOPs. Specifically, $B_i=65536$ tokens are evenly distributed across $N$ experts, with $H=D=8192$.}
\label{fig:group-gemm-benchmark}
\end{figure}
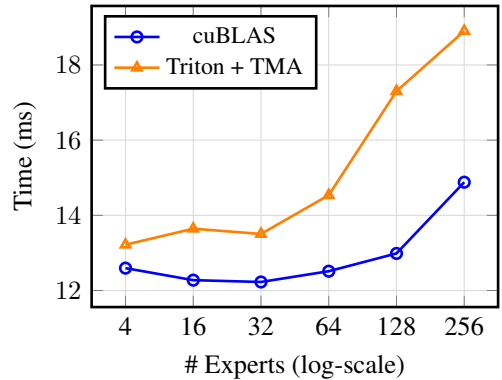

\Cref{fig:group-gemm-benchmark} shows the compute time between a naive for-loop of GEMMs using cuBLAS implementation vs a fused optimized Grouped-GEMM kernel written in Triton, with adoption of Tensor Memory Accelerator (TMA). The cuBLAS version launches $N$ GPU kernel launches, causing high overhead, while the Triton version launches only one. However, as shown, the cuBLAS version still outperforms Triton counterpart because each cuBLAS GEMM kernel is hardware-specific and highly optimized at architecture level, while the Triton version is a generic implementation. In addition, despite all computations have the same FLOPs, the compute time dramatically increases the more experts are present. Therefore, it is better to compute a few giant GEMMs with few experts than to compute many tiny GEMMs with many experts. Both expert parallelism and our method (\ourmethod) leverage this principle by spreading expert weights across EP ranks, allowing each rank to compute only a handful of experts.

\subsection{Number of Experts}\label{sec:app:num_experts}

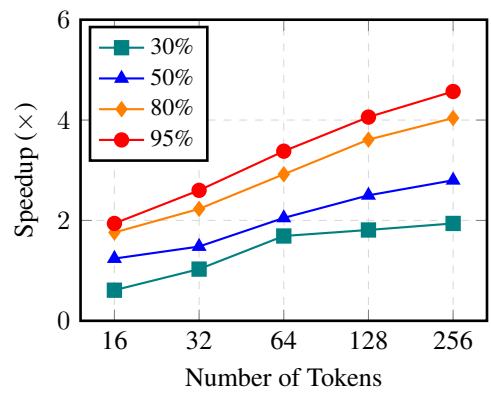
\begin{figure}[t]
\centering
\begin{tikzpicture}
\begin{axis}[
    width=0.5\columnwidth,
    height=0.4\columnwidth,
    xlabel={Number of Tokens},
    ylabel={Speedup ($\times$)},
    xmode=log,
    log basis x=2,
    xtick={16, 32, 64, 128, 256},
    xticklabels={16, 32, 64, 128, 256},
    ymin=0,
    ymax=6,
    legend style={at={(0.02,0.98)}, anchor=north west, font=\small},
    grid=major,
    grid style={dashed, gray!30},
    mark size=2.5pt,
    line width=1pt,
]
\addplot[color=teal, mark=square*, thick] coordinates {
    (16, 0.61) (32, 1.03) (64, 1.69) (128, 1.81) (256, 1.94)
};
\addplot[color=blue, mark=triangle*, thick] coordinates {
    (16, 1.24) (32, 1.48) (64, 2.05) (128, 2.5) (256, 2.8)
};
\addplot[color=orange, mark=diamond*, thick] coordinates {
    (16, 1.76) (32, 2.23) (64, 2.92) (128, 3.61) (256, 4.04)
};
\addplot[color=red, mark=*, thick] coordinates {
    (16, 1.94) (32, 2.6) (64, 3.38) (128, 4.06) (256, 4.57)
};

\legend{30\%, 50\%, 80\%, 95\%}
\end{axis}
\end{tikzpicture}
\caption{Speedup of \ourmethod over standard EP as a function of number of experts ($N$) with 4 imbalanced experts. }
\label{fig:speedup_vs_experts}
\end{figure}

Similar to the trends observe with batch size $B_i$ and hidden sizes $H,D$, \Cref{fig:speedup_vs_experts} shows that \ourmethod is more efficient and exhibits greater speedups when the number of experts ($N$) in the MoE layer increases.

\end{document}